\documentclass[runningheads]{llncs}
\usepackage[T1]{fontenc}
\usepackage{amsmath} 
\usepackage{amssymb}  
\usepackage{graphicx}
\usepackage{caption}
\usepackage{subcaption}
\usepackage{multirow}
\usepackage{amsfonts}
\usepackage{tabularx}
\usepackage{amsfonts}
\usepackage{algorithm,algpseudocode}
\usepackage{bm}
\usepackage{enumerate}
\usepackage{verbatim}
\usepackage{float}
\usepackage{siunitx}
\usepackage{mdframed}
\usepackage{adjustbox}
\usepackage{authblk}
\usepackage{color}
\usepackage{verbatim}
\usepackage{float}
\usepackage{siunitx}
\usepackage{mdframed}
\usepackage{booktabs}
\makeatletter
\let\NAT@parse\undefined
\makeatother
\usepackage{hyperref}
\usepackage{cleveref}

\usepackage{svg}
\usepackage{graphicx}
\usepackage{caption}

\begin{document}
\title{Adaptive Keyframe Selection for Scalable 3D Scene Reconstruction in Dynamic Environments}

\author{Raman Jha\orcidID{0000-0001-8392-8792} \and
Yang Zhou\orcidID{0000-0002-2001-427X} \and
Giuseppe Loianno\orcidID{0000-0002-3263-5401}}

\institute{ Department of Electrical and Computer Engineering, New York University, Brooklyn, NY 11201, USA \\
\email{\{ramanjha, yangzhou\}@nyu.edu}
\and
Department of Electrical Engineering and Computer Sciences, University of California, Berkeley, CA 94720, USA \\
\email{loiannog@eecs.berkeley.edu}}

\maketitle

\begin{abstract}

In this paper, we propose an adaptive keyframe selection method for improved 3D scene reconstruction in dynamic environments using RGB-D sensors. The proposed method integrates two complementary modules: an error-based selection module utilizing photometric and structural similarity (SSIM) errors derived from depth-based warping, and a momentum-based update module that dynamically adjusts keyframe selection thresholds according to scene motion dynamics. By dynamically curating the most informative frames, our approach addresses a key data bottleneck in real-time perception. This allows for the creation of high-quality 3D world representations from a compressed data stream, a critical step towards scalable robot learning and deployment in complex, dynamic environments. Experimental results demonstrate significant improvements over traditional static keyframe selection strategies, such as fixed temporal intervals or uniform frame skipping. These findings highlight a meaningful advancement toward adaptive perception systems that can dynamically respond to complex and evolving visual scenes. We evaluate our proposed adaptive keyframe selection module on two recent state-of-the-art 3D reconstruction networks, Spann3r and CUT3R, and observe consistent improvements in reconstruction quality across both frameworks. Furthermore, an extensive ablation study confirms the effectiveness of each individual component in our method, underlining their contribution to the overall performance gains.

\end{abstract}

\quad

\textbf{Codes}: \href{https://github.com/jhakrraman/Adaptive_Keyframe_Selection}{Adaptive Keyframe Selection}


\section{Introduction}
The pursuit of autonomous robotic systems capable of operating robustly in complex, human-centric environments is a foundational goal in robotics research. A key prerequisite for such autonomy is the ability to perceive, understand, and construct accurate 3D models of the surrounding world in real-time. Historically, perception in robotics evolved from sparse geometric methods, such as those used in early Simultaneous Localization and Mapping (SLAM), which prioritized efficiency for localization tasks \cite{mur2015orb}. However, with the increasing demand for robots to perform sophisticated interaction and navigation, the focus has shifted towards creating dense, high-fidelity 3D reconstructions that capture both the geometry and appearance of the environment. This shift has been accompanied by a dramatic increase in the volume and velocity of sensor data. A modern robotic platform, equipped with multiple high-resolution cameras, can generate several gigabytes of data per minute, creating a formidable computational challenge known as the data bottleneck \cite{9211790,11045277}.

While many state-of-the-art 3D reconstruction methods \cite{wang2024dust3r,wang20243d,wang2025continuous} have demonstrated remarkable progress in building geometric models, their practical utility in robotics is often constrained by their approach to data ingestion. Many of these systems rely on processing a fixed or uniformly sampled subset of frames from a video stream \cite{zhang2024monst3r}. This static approach is fundamentally inefficient, as it operates on the flawed assumption that all moments in time provide equal informational value. For instance, a robot waiting for an elevator door to open perceives a largely static scene where most incoming frames are redundant. Conversely, a robot navigating a crowded hallway experiences rapid and significant changes that require dense temporal sampling to capture accurately. A static selection strategy is ill-equipped to handle this variability; it either squanders computational resources on redundant data in low-activity scenarios or fails to capture critical transient events during high-activity periods, thereby limiting the overall scalability and reactivity of the perception system \cite{wang2025continuous,zhang2024monst3r}. Addressing this inefficiency is a classic problem at the intersection of Computer Vision, which provides the tools for scene analysis; Intelligent Systems, which enable adaptive and data-driven decision-making; and Robotics, which demands efficient and robust perception in complex, real-world environments.

While classical SLAM systems like ORB-SLAM and DSO utilize dynamic keyframe insertion strategies based on tracking quality or parallax, these are primarily optimized for localization stability. In contrast, dense reconstruction often defaults to static temporal intervals or uniform skipping to manage data rates. We distinguish our work by focusing on reconstruction quality rather than tracking. Keyframe selection can be broadly categorized into static (fixed interval) 
\cite{chen2022tensorf,wang2024dust3r,wang20243d,zhang2024monst3r,wang2025continuous}, heuristic (motion-based) \cite{8698793}, geometric (parallax/overlap) \cite{chakraborty2015adaptive}, and appearance-based (photometric) methods [ours]. Our approach bridges these by combining geometric warping with appearance-based metrics to optimize the density of the 3D representation.

To this end, we introduce an adaptive keyframe selection framework that functions as an intelligent, content-aware perception front-end for robotic systems. Rather than treating the video stream as a monolithic data source, our approach dynamically analyzes the information content of each frame, prioritizing those that introduce significant new visual information. By quantifying scene changes through a robust, hybrid error metric, our method intelligently filters the incoming data stream, focusing the finite computational resources of the robotic platform on the frames that matter most. This content-aware selection enables the creation of higher-fidelity 3D world representations from a compressed and more informative subset of the available data. We propose that this is a critical step towards building scalable perception pipelines that are both accurate and efficient enough for real-world deployment. In summary, our key contributions are:
\begin{enumerate}
    \item We propose an adaptive keyframe selection method to address the data bottleneck in real-time world understanding for dynamic and cluttered environments.
    \item Our approach utilizes a hybrid of photometric and SSIM errors, coupled with a momentum-aware dynamic threshold, to intelligently adapt to scene complexity and motion.
    \item We demonstrate through extensive experiments that our method significantly improves reconstruction quality and efficiency when integrated into state-of-the-art networks, validating its role as a scalable perception module.
\end{enumerate}
This paper is organized as follows: Section II provides a comprehensive review of the literature on keyframe selection strategies. Section III details the architecture and mathematical formulation of our proposed method. Section IV presents a thorough experimental evaluation, including quantitative, qualitative, and ablation studies. Finally, Section V discusses the implications of our work and outlines promising directions for future research.

\section{Related Works}
The problem of selecting an optimal subset of frames from a video sequence is a long-standing challenge in computer vision and robotics. The quality of this selection directly impacts the trade-off between computational cost, memory footprint, and the fidelity of downstream tasks such as SLAM or 3D reconstruction. In this section, we provide a structured overview of existing keyframe selection strategies, categorizing them into static, classical adaptive, and modern appearance-based approaches.

\subsection{Static and Heuristic-Based Selection Strategies}
The most straightforward approach to keyframe selection involves the use of static, predefined rules. These methods are computationally inexpensive and easy to implement, making them a common choice in many systems. The simplest strategy is uniform temporal sampling, where a frame is selected every $n$ seconds or frames. A slightly more sophisticated heuristic involves spatial sampling, where a new keyframe is selected after the camera has traveled a certain distance or rotated by a certain angle. While effective in scenarios with uniform motion, these methods are non-adaptive and perform poorly in dynamic environments with variable complexity. Seminal SLAM frameworks like ORB-SLAM \cite{mur2015orb} and ElasticFusion \cite{whelan2016elasticfusion}, MonoSLAM \cite{davison2007monoslam} employ more advanced insertion policies, but these are still based on predefined heuristics that are not optimized for scenarios with highly variable motion dynamics. Many recent and powerful scene reconstruction methods have also adopted static selection policies for simplicity, thereby restricting their robustness. These include TensoRF \cite{chen2022tensorf}, DUST3R \cite{wang2024dust3r}, Spann3r \cite{wang20243d}, Monst3R \cite{zhang2024monst3r}, and CUT3R \cite{wang2025continuous}. The reliance of these state-of-the-art methods on static keyframe strategies fundamentally limits their performance and efficiency in evolving, real-world scenes.

\subsection{Adaptive Keyframe Selection}
To overcome the rigidity of static methods, a significant body of research has focused on adaptive keyframe selection. These techniques aim to select frames based on the content of the scene and the dynamics of the camera's motion, leading to a more efficient and robust selection process. Such approaches have demonstrated clear advantages across various domains. In visual SLAM, adaptive methods \cite{8698793,zu2024adaptive} that select keyframes based on tracking quality or geometric measures like parallax have been shown to improve robustness under complex camera poses. One notable work by Zu et al. \cite{zu2024adaptive} proposed a dynamic selection method for SLAM using a fuzzy inference mechanism based on pose uncertainty; however, its evaluation was primarily focused on localization accuracy in static scenes rather than selection efficiency for dense reconstruction. In other domains, such as video summarization and understanding \cite{Tang_2025_CVPR,chakraborty2015adaptive}, as well as online NeRF-based construction \cite{wilkinson2024adaptive}, adaptive strategies have consistently and significantly outperformed their static counterparts. These works collectively demonstrate the clear benefits of allowing the system to concentrate on the most informative frames, thereby capturing richer detail with fewer computational resources.

Our work builds upon the principles of adaptive selection but distinguishes itself by introducing a novel combination of techniques specifically tailored for modern, transformer-based 3D reconstruction in dynamic environments. Our primary contribution is a framework that intelligently modulates the keyframe selection rate based on a nuanced understanding of visual change.
Another SLAM method, like Direct Sparse Odometry (DSO) \cite{engel2016directsparseodometry}, uses a multi-stage approach: dense keyframes are first created at a high rate, guided by mean optical flow, rotation-compensated flow, and exposure-change metrics. Redundant keyframes are later removed to maintain efficiency. Due to the computational complexity and challenges of optical flow in SLAM systems, the authors employ sparse optical flow information in this work. Firstly, we propose a hybrid error metric that is novel in its fusion of photometric and structural similarity (SSIM) errors for this task. Unlike methods that depend on single metrics or geometric uncertainty \cite{zu2024adaptive}, our approach is robust to both textureless regions (where SSIM excels) and fine-grained geometric changes (captured by photometric error). Secondly, we introduce a momentum mechanism, which is a twofold control strategy for the selection threshold. While the concept of dynamic thresholding has been explored \cite{zhu2022nice}, we uniquely combine it with a post-selection decay factor, which prevents the redundant and bursty selection of frames during periods of continuous high-motion. Finally, we are the first to design and rigorously validate such an adaptive strategy as a plug-in module for state-of-the-art reconstruction networks like Spann3r \cite{wang20243d} and CUT3R \cite{wang2025continuous}, demonstrating significant gains in both efficiency and reconstruction quality on challenging dynamic scene datasets.

\section{Proposed Method}

\subsection{Methodology Overview}
To address the challenge of processing high-bandwidth video streams for real-time 3D reconstruction, we propose an adaptive keyframe selection framework. Our method embodies the synergy between our target fields: it leverages core Computer Vision techniques to generate an error signal, which is then processed by an Intelligent Systems module to make adaptive decisions for a real-time Robotics application. The guiding principle is to discard temporally redundant frames and select only those that introduce significant new information about the scene's geometry or appearance. This is achieved through an online, frame-by-frame decision process that operates as a content-aware filter, ensuring that the downstream reconstruction network receives a compact yet information-rich set of images.

\begin{figure}[htbp]
    \centering
    \includegraphics[width=\textwidth]{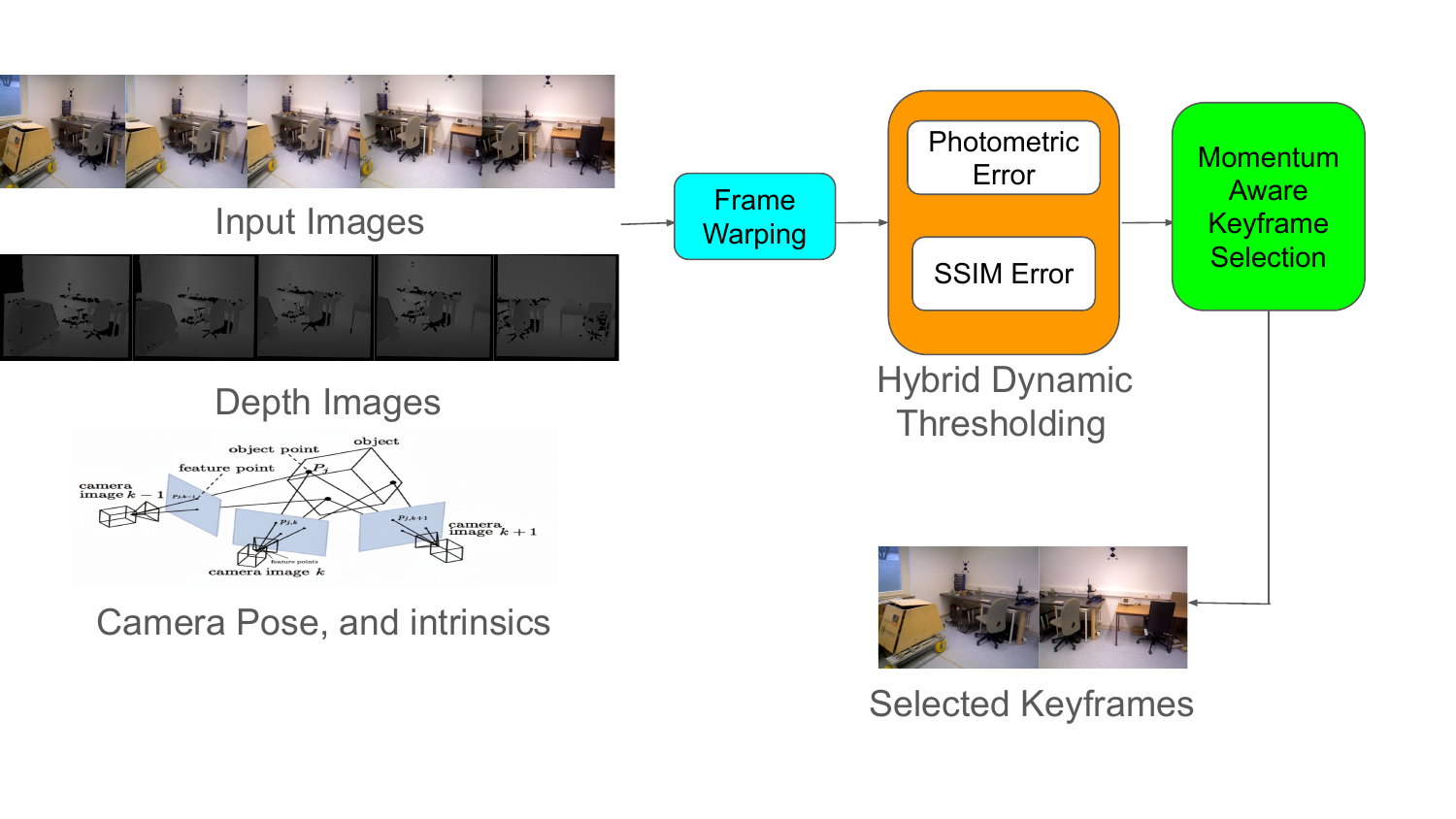}
    \caption{The proposed architecture incorporates an Adaptive Keyframe Selection module that selects informative RGB-D frames based on photometric and structural differences. This content-aware strategy reduces redundancy and improves reconstruction by focusing on keyframes that capture meaningful scene changes. It enables more efficient and accurate 3D reconstruction, especially in dynamic environments.}
    \label{fig:architecture}
\end{figure}

The proposed pipeline, detailed in Algorithm~\ref{alg:adaptive_selection}, follows a clear, sequential process for each incoming frame $f_t$:
\begin{enumerate}
    \item \textbf{Error Computation:} We first compute a \textit{hybrid error score}, $e_t$, that quantifies the perceptual visual change between the current frame $f_t$ and the most recent keyframe, $f_{\text{last\_kf}}$.
    \item \textbf{Dynamic Thresholding:} This error score is then compared against a \textit{dynamic threshold}, $\theta_t$, which adapts in real-time to the scene's level of activity, or "momentum," based on the statistics of recent error scores.
    \item \textbf{Selection Decision:} If the error $e_t$ exceeds the threshold $\theta_t$, the frame is selected as a new keyframe; otherwise, it is discarded.
\end{enumerate}
This closed-loop process allows the system to be highly selective in low-activity scenes while becoming more responsive when significant changes occur. This strikes a crucial balance between the fidelity of the final 3D reconstruction and the computational efficiency required for real-time robotic operation.

\subsection{Hybrid Error Metric for Change Detection}
A robust keyframe selection strategy hinges on accurately measuring the perceptual difference between frames. Relying on a single error metric can be brittle; for instance, simple photometric error is highly sensitive to global illumination changes, while purely structural metrics may miss fine-grained texture changes that are critical for dense reconstruction. To overcome these individual weaknesses, we employ a hybrid error metric that combines both photometric and structural dissimilarity, providing a more comprehensive and robust measure of scene change.

Given a new incoming RGB-D frame $I_t$ at timestep $t$ and the last selected keyframe $I_k$, we first warp $I_k$ into the camera view of $I_t$. This is achieved by projecting the 3D points derived from the depth map of $I_k$ into the image plane of $I_t$ using the known camera poses and intrinsic parameters. This process yields a warped image $\hat{I}_k$ and a validity mask $\mathcal{M}$ indicating the pixels in $I_t$ that have a valid correspondence in $I_k$. The error is then computed only over this set of valid pixels.

\subsubsection{Photometric Error}
The photometric error, $e_{\text{photo}}$, quantifies the difference in pixel intensities between two images. It is highly effective at capturing changes in texture and other fine details. We calculate this as the Mean Absolute Error (L1 norm) over all valid pixels:
\begin{equation}
e_{\text{photo}}(I_t, \hat{I}_k) = \frac{1}{|\mathcal{M}|} \sum_{p \in \mathcal{M}} |I_t(p) - \hat{I}_k(p)|,
\label{eq:photometric}
\end{equation}
where $p$ represents a pixel coordinate in the valid mask $\mathcal{M}$, and $I(p)$ is the pixel intensity. We specifically choose the L1 norm over the L2 norm (Mean Squared Error) for its increased robustness to outliers, such as sensor noise or small occlusions, which might otherwise disproportionately influence the error score. The valid mask $\mathcal{M}$ identifies pixels with successful depth-based reprojection between frames. We observe that as the overlap decreases, common in highly dynamic sequences, the mask size shrinks, naturally leading to a higher error signal that triggers keyframe insertion. Conversely, in static datasets with high overlap, the error remains stable. This mechanism ensures that the system is most responsive when scene coverage is at risk.

\subsubsection{Structural Similarity (SSIM) Error}
To enhance robustness against non-structural variations like global illumination and exposure changes, we incorporate the Structural Similarity Index Measure (SSIM) \cite{wang2004image}. SSIM is a perception-based model that compares local patterns of pixel intensities by evaluating three distinct components: luminance, contrast, and structure. By focusing on these perceptual characteristics, it provides a measure of similarity that aligns more closely with human visual perception and is less sensitive to absolute intensity values. We define the SSIM error, $e_{\text{ssim}}$, as the structural dissimilarity over the valid region $\mathcal{M}$:
\begin{equation}
e_{\text{ssim}}(I_t, \hat{I}_k) = 1 - \text{SSIM}(I_t, \hat{I}_k).
\label{eq:ssim}
\end{equation}

\subsubsection{Combined Error Score}
The final error score for the current frame, $e_t$, is a weighted linear combination of the photometric and SSIM errors:
\begin{equation}
e_t = \alpha \cdot e_{\text{photo}} + \beta \cdot e_{\text{ssim}}.
\label{eq:combined}
\end{equation}
where $\alpha$ and $\beta$ are hyperparameters that balance the two metrics. The weighting allows us to tune the sensitivity of the error metric. In our experiments, we set $\alpha=0.7$ and $\beta=0.3$, prioritizing the detection of fine-grained photometric detail, which is crucial for high-fidelity dense reconstruction, while still benefiting from the stability and robustness of the SSIM metric.

\begin{algorithm}[h]
\caption{Hybrid Error Computation}
\label{alg:error_computation}
\begin{algorithmic}[1]
\Function{ComputeHybridError}{$I_t, I_k, D_k, \text{Pose}_t, \text{Pose}_k, \text{Intrinsics}$}
    \State $\hat{I}_k, \mathcal{M} \leftarrow \texttt{WarpFrame}(I_k, D_k, \text{Pose}_k, \text{Pose}_t, \text{Intrinsics})$
    \State $e_{\text{photo}} \leftarrow \texttt{CalculateL1Error}(I_t, \hat{I}_k, \mathcal{M})$
    \State $e_{\text{ssim}} \leftarrow \texttt{CalculateSSIMError}(I_t, \hat{I}_k, \mathcal{M})$
    \State $e_t \leftarrow \alpha \cdot e_{\text{photo}} + \beta \cdot e_{\text{ssim}}$
    \State \textbf{return} $e_t$
\EndFunction
\end{algorithmic}
\end{algorithm}

\subsection{Momentum-Aware Dynamic Thresholding}
Using a fixed error threshold for keyframe selection is suboptimal, as it fails to adapt to the natural ebbs and flows of activity in a dynamic scene. A low threshold would be too permissive in a high-action scene, selecting many redundant frames, while a high threshold would be too restrictive in a slow-moving scene, potentially missing important subtle changes. To address this, we introduce a dynamic threshold, $\theta_t$, that adapts to the scene's momentum. We quantify this concept by treating the recent history of frame-to-frame errors as a statistical signal representing the scene's current rate of change.

The core selection rule is therefore a simple comparison: a frame $f_t$ is selected as a new keyframe if its computed error $e_t$ exceeds the dynamic threshold $\theta_t$. The intelligence of the system lies in how this threshold $\theta_t$ is computed and updated over time.

\subsubsection{Adaptive Threshold Computation}
The threshold $\theta_t$ is computed from a sliding window of the $W$ most recent error scores. This provides a localized statistical model of the scene's recent activity. We first calculate the moving average $\mu_t$ and standard deviation $\sigma_t$ of these errors:
\begin{equation}
\mu_t = \frac{1}{W} \sum_{i=t-W+1}^{t} e_i,
\end{equation}
\begin{equation}
\sigma_{t} = \sqrt{\frac{1}{W}\sum_{i=t-W+1}^{t}(e_{i}-\mu_{t})^{2}}.
\end{equation}
The moving average $\mu_t$ represents the current baseline of visual change, while the standard deviation $\sigma_t$ quantifies its volatility. The dynamic threshold is then set as an upper control limit based on these statistics, ensuring it never falls below a predefined base threshold $\theta_0$:
\begin{equation}
\theta_t = \max(\theta_0, \mu_t + k \cdot \sigma_t).
\label{eq:threshold}
\end{equation}
Here, $k$ is a sensitivity parameter. This formulation allows the threshold to automatically rise during periods of high and variable motion and fall during periods of calm, ensuring that only statistically significant changes trigger a keyframe selection.

\subsubsection{Post-Selection Threshold Decay}
During a period of sustained high motion (e.g., rapid camera rotation), the error can remain consistently above even an elevated adaptive threshold. This can lead to a burst of closely-spaced keyframes that, while individually significant, are collectively redundant. To prevent this and promote temporal diversity, we apply a decay factor, $\gamma$, to the threshold immediately after a new keyframe is selected:
\begin{equation}
\theta_{\text{new}} = \gamma \cdot \theta_t, \quad \text{where } 0 < \gamma < 1.
\label{eq:decay}
\end{equation}
This momentary reduction of the threshold introduces a "refractory period," a concept analogous to mechanisms that prevent over-firing in biological neural systems. This makes it less likely for the very next frame to also be selected, ensuring that the chosen keyframes are more meaningfully spaced over time.

\begin{algorithm}[t]
\caption{Momentum-Aware Adaptive Keyframe Selection}
\label{alg:adaptive_selection}
\begin{algorithmic}[1]
\Require Sequence of frames $\mathcal{F} = \{f_1, f_2, \dots, f_n\}$
\Require Base threshold $\theta_0$, window size $W$
\Require Sensitivity $k$, decay factor $\gamma$
\Ensure Selected keyframes $\mathcal{K}$

\State Initialize $\mathcal{K} \leftarrow \{f_1\}$, $f_{\text{last\_kf}} \leftarrow f_1$
\State Initialize error history $\mathcal{E} \leftarrow []$, $\theta \leftarrow \theta_0$

\For{$t = 2$ to $n$}
    \State $e_t \leftarrow$ \texttt{ComputeHybridError}$(f_t, f_{\text{last\_kf}})$ \Comment{See Algorithm~\ref{alg:error_computation}}
    \State Append $e_t$ to $\mathcal{E}$
    
    \If{$|\mathcal{E}| \geq W$} \Comment{If window is full}
        \State $\mu_t \leftarrow \text{mean of last } W \text{ errors in } \mathcal{E}$
        \State $\sigma_t \leftarrow \text{std of last } W \text{ errors in } \mathcal{E}$
        \State $\theta \leftarrow \max(\theta_0, \mu_t + k \cdot \sigma_t)$ \Comment{See Eq.~\ref{eq:threshold}}
    \Else \Comment{Warm-up phase for initial frames}
        \State $\theta \leftarrow \theta_0 \cdot \frac{t}{W} + \theta_{init} \cdot (1 - \frac{t}{W})$
    \EndIf

    \If{$e_t > \theta$}
        \State Append $f_t$ to $\mathcal{K}$
        \State $f_{\text{last\_kf}} \leftarrow f_t$
        \State $\theta \leftarrow \gamma \cdot \theta$ \Comment{Apply decay, see Eq.~\ref{eq:decay}}
    \EndIf
\EndFor
\State \textbf{return} $\mathcal{K}$
\end{algorithmic}
\end{algorithm}

\subsection{Hyperparameter Analysis}
The core logic of our error computation is summarized in Algorithm~\ref{alg:error_computation}. In our experiments, these were set as follows based on empirical evaluation on a held-out validation dataset:
\begin{itemize}
    \item \textbf{Error Weights ($\alpha, \beta$):} Set to $\alpha=0.7$ and $\beta=0.3$ to prioritize the capture of fine-grained detail. The behavior of our framework is governed by a set of key hyperparameters. The parameters $\alpha$ and $\beta$ in Eq. \eqref{eq:combined} are treated independently rather than as a single mixing parameter to allow fine-grained control over the system's sensitivity to sudden lighting changes (captured by SSIM) versus raw pixel intensity shifts.
    \item \textbf{Window Size ($W$):} A window size of $W=5$ frames was used. This provides a balance between statistical stability and reactivity to sudden changes in scene dynamics.  We employ a small window size $W=5$ for the moving average to prioritize high responsiveness to rapid motion, which a standard Exponential Moving Average (EMA) might over-smooth. 
    \item \textbf{Sensitivity ($k$):} The sensitivity factor was set to $k=1.5$. A higher value makes the selection more conservative, while a lower value makes it more permissive. This value was found to be effective at filtering minor motion noise without suppressing genuinely novel frames.
    \item \textbf{Decay Factor ($\gamma$):} The decay factor was set to $\gamma=0.95$. This value provides a sufficient refractory period to prevent burst selections without making the system insensitive after a selection event.
    \item \textbf{Base Threshold ($\theta_0$):} A small base threshold was set to ensure a minimum level of selectivity even in completely static scenes. The base threshold $\theta_0$ was determined through a grid search on the 'Bonn Dynamic' validation subset, specifically targeting a balance between reconstruction accuracy and frame count.
\end{itemize}

\section{Experiments}
To rigorously evaluate the performance and adaptability of our proposed keyframe selection framework, we conducted a comprehensive set of experiments across a diverse range of public datasets. This section details our experimental methodology, including the datasets, evaluation metrics, and baselines used for comparison. We aim to answer three primary research questions: 1) Does our adaptive method improve reconstruction quality and efficiency over static selection strategies in dynamic environments? 2) How does our method perform in static environments compared to established heuristics? 3) What is the contribution of each component within our proposed framework? Our adaptive selection module is designed to be a lightweight front-end, capable of being integrated into various modern reconstruction networks to enhance their adaptability and performance on diverse robotic platforms.

\subsubsection{Integration with Host Networks}
To ensure a fair and direct comparison, we integrated our adaptive keyframe selection module into two state-of-the-art 3D reconstruction networks: Spann3r \cite{wang20243d} and CUT3R \cite{wang2025continuous}. For both networks, we used the official, publicly available source code and pretrained model weights. The only modification made was replacing their default static keyframe selection logic with our proposed module. All other aspects of the networks, including their internal architectures, training protocols, and hyperparameters, were kept identical to their original implementations. This isolates the impact of the keyframe selection strategy on the final reconstruction performance.

\subsection{Datasets and Evaluation Metrics}
To demonstrate the generalization capability of our system, we evaluate on five challenging datasets covering a wide spectrum of robotic scenarios. We use three static datasets: \textbf{7Scenes}\cite{glocker2013real,shotton2013scene}, which contains challenging indoor trajectories with textureless regions; \textbf{NRGBD} \cite{Azinovic_2022_CVPR}, an RGB-D odometry dataset; and \textbf{DTU} \cite{jensen2014large}, a standard benchmark for multi-view stereo. We also use two dynamic datasets: MPI Sintel \cite{Butler:ECCV:2012}, a synthetic dataset with complex non-rigid motion, and \textbf{Bonn} \cite{palazzolo2019iros}, a real-world dataset featuring significant human motion.

We evaluate reconstruction quality using a suite of standard metrics as in prior works \cite{wang2024dust3r,wang20243d,zhang2024monst3r,wang2025continuous}: \textbf{Accuracy} (Acc↓) and \textbf{Completion} (Comp↓) measure the distance between the predicted and ground truth point clouds, \textbf{Normal Consistency} (NC↑) assesses surface smoothness, and \textbf{Chamfer Distance} (CD↓) captures overall shape similarity. To specifically quantify selection efficiency, we introduce the \textbf{Keyframe Compression Ratio (KFCR↑)}. This metric represents the percentage of frames discarded from the original sequence, where a higher value indicates greater efficiency and compression. Since our primary contribution is the adaptive selection of the most informative frames for geometry, the KFCR and reconstruction quality serve as the most relevant indicators of performance. We prioritize surface metrics such as Accuracy, Completeness, and Chamfer Distance over pose metrics (ATE/RPE). 

\subsection{Quantitative Analysis}
To provide a comprehensive performance benchmark, we conducted a rigorous quantitative comparison of our adaptive method against the static keyframe selection strategy native to our two host networks, Spann3r \cite{wang20243d} and CUT3R \cite{wang2025continuous}. The analysis is divided into performance on static and dynamic datasets to clearly delineate the method's behavior under different conditions.

\subsubsection{Performance on Static Datasets}
The results on the 7Scenes \cite{glocker2013real,shotton2013scene}, NRGBD \cite{Azinovic_2022_CVPR}, and DTU \cite{jensen2014large}datasets are presented in Table~\ref{static1} and Table~\ref{static2}. On these datasets, which lack significant dynamic objects, our method performs competitively and often demonstrates slight improvements. For instance, when integrated with CUT3R on 7Scenes \cite{glocker2013real,shotton2013scene}, our method improves on all metrics, indicating a better selection of geometrically informative viewpoints.

\begin{table}[htbp]
\centering
\begin{adjustbox}{max width=\textwidth}
\begin{tabular}{lccccc}
\toprule
\textbf{Methods} & \textbf{Acc↓} & \textbf{Comp↓} & \textbf{Chamfer dis↓} & \textbf{NC↑} & \textbf{KFCR↑} \\
\midrule
\multicolumn{6}{c}{\textbf{7 Scenes}} \\
\midrule
Spann3r & 0.0314 & 0.0206 & 0.0260 & \textbf{0.6756} & \textbf{95.00} \\
Adaptive & \textbf{0.0292} & \textbf{0.0190} & \textbf{0.0241} & 0.6747 & 94.52 \\
\midrule
\multicolumn{6}{c}{\textbf{NRGBD}} \\
\midrule
Spann3r & 0.0805 & 0.0326 & 0.0566 & \textbf{0.7779} & \textbf{97.50} \\
Adaptive & \textbf{0.0781} & \textbf{0.0321} & \textbf{0.0551} & 0.7184 & 94.88 \\
\midrule
\multicolumn{6}{c}{\textbf{DTU}} \\
\midrule
Spann3r & \textbf{4.930} & \textbf{2.754} & 4.241 & 0.741 & 80.00 \\
Adaptive & 5.241 & 3.042 & \textbf{3.783} & \textbf{0.764} & \textbf{82.70} \\
\bottomrule
\end{tabular}
\end{adjustbox}
\caption{Comparison with \textbf{Spann3r} on static datasets.}
\label{static1}
\end{table}

On datasets with rich texture and ideal conditions like DTU, the performance of the static baseline is already very strong. In these cases, our adaptive method may trade a minor decrease in one metric (e.g., Accuracy) for a significant gain in another (e.g., Chamfer Distance and NC), suggesting that our selection prioritizes overall geometric fidelity. The minor fluctuations in performance highlight a known characteristic of adaptive methods: in highly regular, texture-rich scenes, a simple uniform sampling can be a very effective strategy. However, our method's ability to maintain competitive performance while offering superior efficiency (as shown by the higher KFCR on DTU) underscores its robustness.

\begin{table}[htbp]
\centering
\begin{adjustbox}{max width=\textwidth}
\begin{tabular}{lccccc}
\toprule
\textbf{Methods} & \textbf{Acc↓} & \textbf{Comp↓} & \textbf{Chamfer dis↓} & \textbf{NC↑} & \textbf{KFCR↑} \\
\midrule
\multicolumn{6}{c}{\textbf{7 Scenes}} \\
\midrule
CUT3R & 0.023 & 0.018 & 0.021 & 0.674 & 95.00 \\
Adaptive & \textbf{0.022} & \textbf{0.017} & \textbf{0.020} & \textbf{0.684} & \textbf{95.83} \\
\midrule
\multicolumn{6}{c}{\textbf{NRGBD}} \\
\midrule
CUT3R & \textbf{0.730} & 0.038 & \textbf{0.053} & 0.719 & \textbf{97.50} \\
Adaptive & 0.780 & \textbf{0.032} & 0.059 & \textbf{0.787} & 95.88 \\
\midrule
\multicolumn{6}{c}{\textbf{DTU}} \\
\midrule
CUT3R & 3.832 & \textbf{2.140} & \textbf{2.986} & 0.743 & 80.00 \\
Adaptive & \textbf{3.714} & 2.394 & 3.124 & \textbf{0.768} & \textbf{82.70} \\
\bottomrule
\end{tabular}
\end{adjustbox}
\caption{Comparison with \textbf{CUT3R} on static datasets.}
\label{static2}
\end{table}

\subsubsection{Performance on Dynamic Datasets}
The true strength of our adaptive approach becomes evident in dynamic environments. As shown in Table~\ref{dynamic1} and Table~\ref{dynamic2}, our method consistently outperforms the static baseline across the Sintel \cite{Butler:ECCV:2012} and Bonn \cite{palazzolo2019iros} datasets, which are characterized by significant motion and complex scene changes. 

\begin{table}[htbp]
\centering
\begin{adjustbox}{max width=\textwidth}
\begin{tabular}{lccccc}
\toprule
\textbf{Methods} & \textbf{Acc↓} & \textbf{Comp↓} & \textbf{Chamfer dis↓} & \textbf{NC↑} & \textbf{KFCR↑} \\
\midrule
\multicolumn{6}{c}{\textbf{Sintel}} \\
\midrule
Spann3r & 0.1260 & \textbf{0.0252} & 0.0756 & \textbf{0.6842} & 90.00 \\
Adaptive & \textbf{0.1223} & 0.0261 & \textbf{0.0742} & 0.6677 & 90.00 \\
\midrule
\multicolumn{6}{c}{\textbf{Bonn}} \\
\midrule
Spann3r & 0.1614 & 0.0446 & 0.1030 & 0.5406 & 90.00 \\
Adaptive & \textbf{0.1522} & \textbf{0.0408} & \textbf{0.0965} & \textbf{0.5579} & \textbf{90.70} \\
\bottomrule
\end{tabular}
\end{adjustbox}
\caption{Comparison with \textbf{Spann3r} on dynamic datasets.}
\label{dynamic1}
\end{table}

On the challenging real-world Bonn dataset, our method achieves superior results across all reconstruction metrics for both Spann3r and CUT3R, while also using fewer keyframes (higher KFCR). This demonstrates the core advantage of our approach: by intelligently selecting frames that capture meaningful changes, it not only improves efficiency but also leads to a more accurate and complete final reconstruction. The ability to filter redundant information allows the reconstruction network to focus its capacity on modeling the most critical aspects of the evolving scene. A reduction in keyframes leads to a direct linear decrease in back-end optimization memory (G-buffer storage) and computational load for the global mapping process, as fewer frames require fusion into the global model.

\begin{table}[htbp]
\centering
\begin{adjustbox}{max width=\textwidth}
\begin{tabular}{lccccc}
\toprule
\textbf{Methods} & \textbf{Acc↓} & \textbf{Comp↓} & \textbf{Chamfer dis↓} & \textbf{NC↑} & \textbf{KFCR↑} \\
\midrule
\multicolumn{6}{c}{\textbf{Sintel}} \\
\midrule
CUT3R & 0.548 & \textbf{0.014} & 0.281 & 0.543 & \textbf{90.00} \\
Adaptive & \textbf{0.410} & 0.026 & \textbf{0.0218} & \textbf{0.574} & 89.42 \\
\midrule
\multicolumn{6}{c}{\textbf{Bonn}} \\
\midrule
CUT3R & 0.2635 & 0.038 & 0.137 & \textbf{0.569} & 90.00 \\
Adaptive & \textbf{0.233} & \textbf{0.034} & \textbf{0.124} & 0.537 & \textbf{91.28} \\
\bottomrule
\end{tabular}
\end{adjustbox}
\caption{Comparison with \textbf{CUT3R} on dynamic datasets.}
\label{dynamic2}
\end{table}

\subsection{Qualitative Analysis}

\begin{figure*}[ht!]
    \centering
    \includegraphics[width=\textwidth, height  = 11cm]{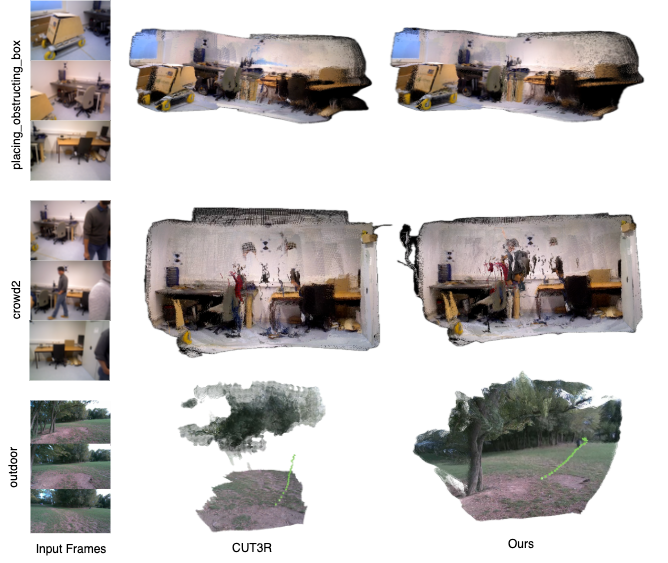}
    \caption{Qualitative Results on the dynamic BONN, and our own collected outdoor dataset. We compare our method with concurrent works, CUT3R. Our method achieves the best qualitative results in complex and cluttered environments.}
    \label{fig:qualitative}
\end{figure*}

Beyond quantitative metrics, we provide a qualitative analysis to inspect the differences in reconstruction quality visually. As illustrated in Figure~\ref{fig:qualitative}, which shows results on challenging sequences from the BONN dataset, and our own collected outdoor scenes, our method produces visibly more complete and detailed 3D models. When integrated with networks such as CUT3R \cite{wang2025continuous}, our adaptive selection helps capture fine details on moving objects and reduces artifacts in the reconstructed geometry. The impact of the module is more significant when evaluated on outdoor scenes, particularly in dynamic environments. This visual evidence supports the quantitative findings and highlights the practical benefits of our approach to real-time robotic perception in dynamic environments.

\subsection{Ablation and Scene-Wise Analysis}
\subsubsection{Ablation Study}
To validate the contribution of each component in our proposed error metric, we performed an ablation study on the dynamic Bonn dataset. The results, summarized in Table~\ref{ablation}, show that both components are crucial for optimal performance. Removing the photometric error forces reliance solely on SSIM, reducing sensitivity to fine-grained textural changes, while excluding SSIM makes the process brittle to illumination changes and prone to missing structural shifts. We further compared our approach to existing adaptive keyframe selection strategies, including inertial- and optimization-based methods. The inertial-based approach \cite{8698793}, though leveraging IMU motion cues, underperforms in accuracy and completeness, indicating that inertial data alone fails to capture appearance-driven structural variations. Similarly, the optimization-based method \cite{chakraborty2015adaptive} achieves slightly better completeness than the inertial-based approach but still lags behind our hybrid strategy, highlighting its tendency to overfit trajectory smoothness at the cost of structural detail.

In contrast, our full model integrating both photometric and SSIM errors achieves the best balance of reconstruction quality and efficiency. It outperforms all baselines in accuracy, Chamfer distance, and normal consistency, while maintaining a competitive keyframe compression ratio. This confirms that the complementary integration of photometric sensitivity and structural robustness yields superior keyframe selection performance compared to prior strategies. The ablation study was conducted on the Bonn dataset as it represents the most diverse and challenging dynamic scenarios in our test suite. However, the performance gains observed in Section 4.2 across other datasets confirm that the threshold update dynamics remain robust and effective across various motion profiles.

\begin{table}[htbp]
\centering
\begin{adjustbox}{max width=\textwidth}
\begin{tabular}{lccccc}
\toprule
\textbf{Methods} & \textbf{Acc↓} & \textbf{Comp↓} & \textbf{Chamfer dis↓} & \textbf{NC↑} & \textbf{KFCR↑} \\
\midrule
w/o Photo & 0.1570 & \textbf{0.0396} & 0.0981 & 0.5478 & 86.80 \\
w/o SSIM & 0.1651 & 0.0500 & 0.1076 & 0.5425 & 93.81 \\
Inertial-based\cite{8698793} & 0.1623 & 0.0550 & 0.1271 & 0.5135 & 81.75 \\
Optimization-based\cite{chakraborty2015adaptive} & 0.1678 & 0.0585 & 0.1205 & 0.5237 & 87.75 \\
Full (Ours) & \textbf{0.1522} & 0.0408 & \textbf{0.0965} & \textbf{0.5579} & \textbf{90.70} \\
\bottomrule
\end{tabular}
\end{adjustbox}
\caption{Ablation study on the components of the hybrid error metric, and other recent adaptive keyframe selection methods evaluated on the BONN dataset.}
\label{ablation}
\end{table}

\begin{figure}[ht!]
    \centering
    \includegraphics[width=\textwidth, width = 12cm]{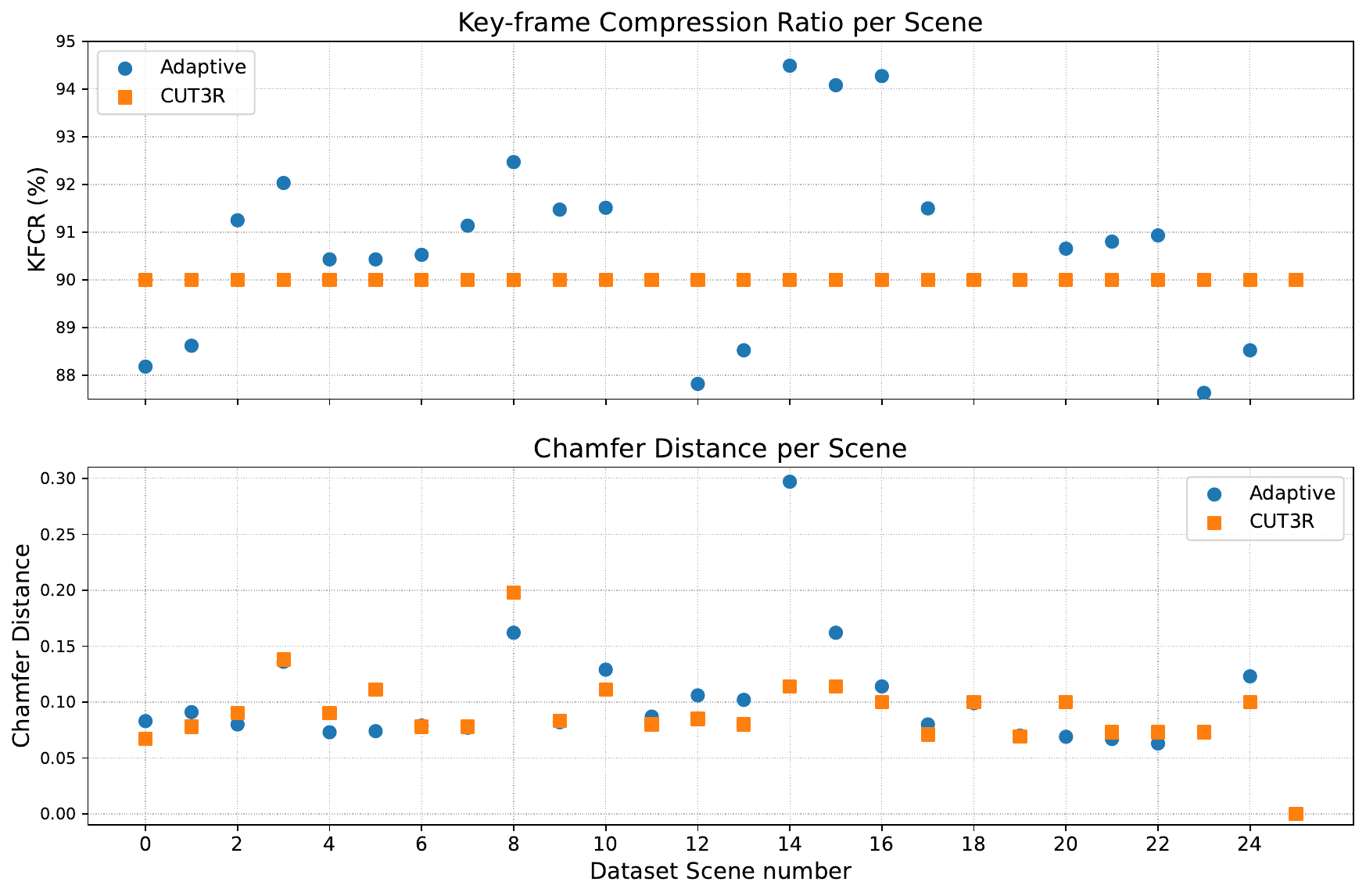}
    \caption{Scene-wise comparison of Keyframe Compression Ratio (KFCR) and Chamfer Distance on the BONN dataset. Our adaptive method (blue) varies its selection rate based on scene complexity, while maintaining competitive reconstruction quality compared to the static CUT3R baseline (orange).}
    \label{fig:comparison}
\end{figure}

\subsubsection{Scene-wise Efficiency and Quality}
Figure~\ref{fig:comparison} presents a scene-by-scene analysis of the BONN dataset, directly visualizing the trade-off between efficiency and quality. The KFCR plot (top) clearly shows that our method's selection rate is not constant; it uses fewer frames in simpler scenes (higher KFCR) and more frames in complex ones (lower KFCR), unlike the fixed-rate baseline. Despite this dynamic variation in frame count, the Chamfer Distance plot (bottom) shows that our method's reconstruction quality remains consistently competitive with, and often better than, the baseline. This demonstrates that our approach successfully preserves reconstruction quality while significantly improving keyframe selection efficiency.

\section{Discussion}
\textbf{Assumptions and Limitations:} Our approach assumes the availability of depth data (RGB-D) for the warping module. While SSIM provides robustness to global lighting changes, the system may still be sensitive to extreme local specularities or saturated regions. Furthermore, in cases of extremely rapid motion where the overlap mask $\mathcal{M}$ becomes negligible, the warping-based error may lose its geometric grounding, though this typically results in a 'fail-safe' trigger of a new keyframe.

\textbf{Modest Gains vs. Efficiency:} While the quantitative improvements in Accuracy and Completeness are modest in some sequences, the primary 'win' of this work is the efficiency gain. We demonstrate that it is possible to achieve parity with or slightly exceed state-of-the-art reconstruction quality while processing significantly fewer frames. This reduction is vital for deploying dense SLAM on resource-constrained robotic platforms.

\section{Conclusion and Future Works}
This paper presented an adaptive keyframe selection method that addresses a critical data bottleneck in real-time 3D perception. By intelligently selecting frames based on scene dynamics, our approach enables the creation of more accurate and detailed 3D reconstructions from significantly fewer data points. The method's effectiveness was demonstrated through consistent improvements in quality and efficiency when integrated into modern reconstruction networks, confirming its value as a practical module for scalable robotic systems operating in dynamic environments. These can then be used to seed generative models to synthesize diverse and physically plausible training scenarios, significantly broadening the data available for training robust robotic policies.

The scientific value of this work lies in the formulation of a momentum-aware thresholding strategy that bridges the gap between simple temporal heuristics and complex, computationally expensive keyframe optimization. Practically, this provides a scalable solution for real-time robotic perception, allowing for high-quality environment modeling under restricted bandwidth and memory constraints. For future work, our work on efficient world understanding serves as a foundational step for several exciting research directions. First, while our hybrid error metric is robust, its reliance on appearance makes it sensitive to challenging illumination conditions \cite{11084466}, which we want to address in the future. Second, we plan to leverage the high-fidelity and compact 3D representations created by our method to address the long-tail data problem in robotics. The curated keyframes and resulting 3D models can serve as grounded "digital twins" of real-world events. 

\begin{credits}
\subsubsection{\ackname} This work was supported by the DARPA YFA Grant D22AP00156-00, the DEVCOM ARL grant SARA W911NF-24-2-0057 grant, and the NSF CPS Grant CNS-2121391.

\end{credits}

\bibliographystyle{IEEEtran} 
\bibliography{root}
\end{document}